%% file: main.tex
\documentclass[10pt,twocolumn,letterpaper]{article}

\usepackage{iccv}
\usepackage{times}
\usepackage{epsfig}
\usepackage{graphicx}
\usepackage{amsmath}
\usepackage{amssymb}
\usepackage{multirow}
\usepackage{url}
\usepackage{graphicx}
\usepackage{algorithm}  
\usepackage{algorithmicx} 
\usepackage{algpseudocode}
\usepackage{adjustbox}
\usepackage{caption}
\usepackage{subfloat}
\usepackage[accsupp]{axessibility} 
\usepackage[pagebackref=true,breaklinks=true,letterpaper=true,colorlinks,bookmarks=false]{hyperref}
\newcommand{\method}{MADAug\xspace}
\newcommand{\trick}{monotonic curriculum\xspace}
\iccvfinalcopy
\def\iccvPaperID{11274} % *** Enter the ICCV Paper ID here

\ificcvfinal\fi

\begin{document}

%%%%%%%%% TITLE
\title{When to Learn What: Model-Adaptive Data Augmentation Curriculum}

\author{%
Chengkai Hou$^{1}$, \
Jieyu Zhang$^{2}$, \
Tianyi Zhou$^{3}$\thanks{Corresponding author} \\
$^1$Jilin university, \ 
$^2$University of Washington, \
$^3$ University of Maryland\\
\texttt{ \small houck20@mails.jlu.edu.cn, jieyuz2@cs.washington.edu, tianyi@umd.edu}\\
}

\maketitle
\ificcvfinal\thispagestyle{empty}\fi
%\iccvfinalcopy

\begin{abstract}
Data augmentation (DA) is widely used to improve the generalization of neural networks by enforcing the invariances and symmetries to pre-defined transformations applied to input data. However, a fixed augmentation policy may have different effects on each sample in different training stages but existing approaches cannot adjust the policy to be adaptive to each sample and the training model. 
In this paper, we propose ``\textbf{M}odel-\textbf{A}daptive \textbf{D}ata \textbf{Aug}mentation (MADAug)'' that jointly trains an augmentation policy network to teach the model ``\textbf{when to learn what}''. 
Unlike previous work, MADAug selects augmentation operators for each input image by a model-adaptive policy varying between training stages, producing a data augmentation curriculum optimized for better generalization. 
In MADAug, we train the policy through a bi-level optimization scheme, which aims to minimize a validation-set loss of a model trained using the policy-produced data augmentations.  
% Data augmentation policies are a common
% way in machine learning
% for improving generalization performance. However, finding the adoptive data augmentation policies requires
% extra knowledge or a computationally demanding search.
% In this paper, we propose to use bilevel optimization to optimize data augmentation policies though the performance of the task model on validation set to address this issue.
% This framework can be used as a general solution to learn the optimal data augmentation policy jointly with many task models like a classifier. 
We conduct an extensive evaluation of MADAug on multiple image classification tasks and network architectures with thorough comparisons to existing DA approaches. MADAug outperforms or is on par with other baselines and exhibits better fairness: it brings improvement to all classes and more to the difficult ones. 
Moreover, MADAug learned policy shows better performance when transferred to fine-grained datasets. In addition, the auto-optimized policy in MADAug gradually introduces increasing perturbations and naturally forms an easy-to-hard curriculum. %\looseness-1
Our code is available at \url{ https://github.com/JackHck/MADAug}.
% Experiment results show that our training method produces an image classification accuracy that is comparable
% to or better than previous methods,including fine-grained image dataets.
% Yet, it does not need an expensive external validation loop
% on the data augmentation hyperparameters.
\end{abstract}
%\vspace{-1.0em}

%% Please note that we have introduced automatic line number generation
%% into the style file for \LaTeXe. This is to help reviewers
%% refer to specific lines of the paper when they make their comments. Please do
%% NOT refer to these line numbers in your paper as they will be removed from the
%% style file for the final version of accepted papers.

\input{01-intro}

\input{02-related.tex}

\input{03-method.tex}

\input{04-exp.tex}

\input{05-discussion.tex}
\clearpage
\bibliographystyle{ieee_fullname}
\bibliography{ref}
\clearpage
\input{06-appendix.tex}

\end{document}

%% file: 01-intro.tex
\section{Introduction}

Data augmentation is a widely used strategy to increase the diversity of training data, which improves the model generalization, especially in image recognition tasks~\cite{krizhevsky2017imagenet,srivastava2015training,hernandez2018data}.  
Unlike previous works that apply manually-designed augmentation operations~\cite{cubuk2018autoaugment,zhang2019adversarial,zhou2021metaaugment,li2020dada,cheung2021AdaAug,lim2019fast}, recent researchers have resorted to searching a data augmentation policy for a target dataset/samples. 
Despite the success of these learnable and dataset-dependent augmentation policies, they are fixed once learned and thus non-adaptive to either different samples or models at different training stages, resulting in biases across different data regions~\cite{balestriero2022effects} or inefficient training. 
% the policies for sample/dataset are fixed during model training. 

%In this paper, we study two aspects to formulate data augmentation polices: \textbf{(1) when to apply data augmentation  during model training; (2) what the data augmentation policies to apply at different training stages}. 

In this paper, we study two fundamental problems towards developing a data-and-model-adaptive data augmentation policy that determines a curriculum of ``when to learn what'' to train a model: 
% propose and study two design principles for a better data augmentation policy: 
\textbf{(1)} \textit{when to apply data augmentation in training?} \textbf{(2)} \textit{what data augmentations should be applied to each training sample at different training stages?}

\begin{figure*}[t]
    \centering
    \includegraphics[width=0.98\linewidth]{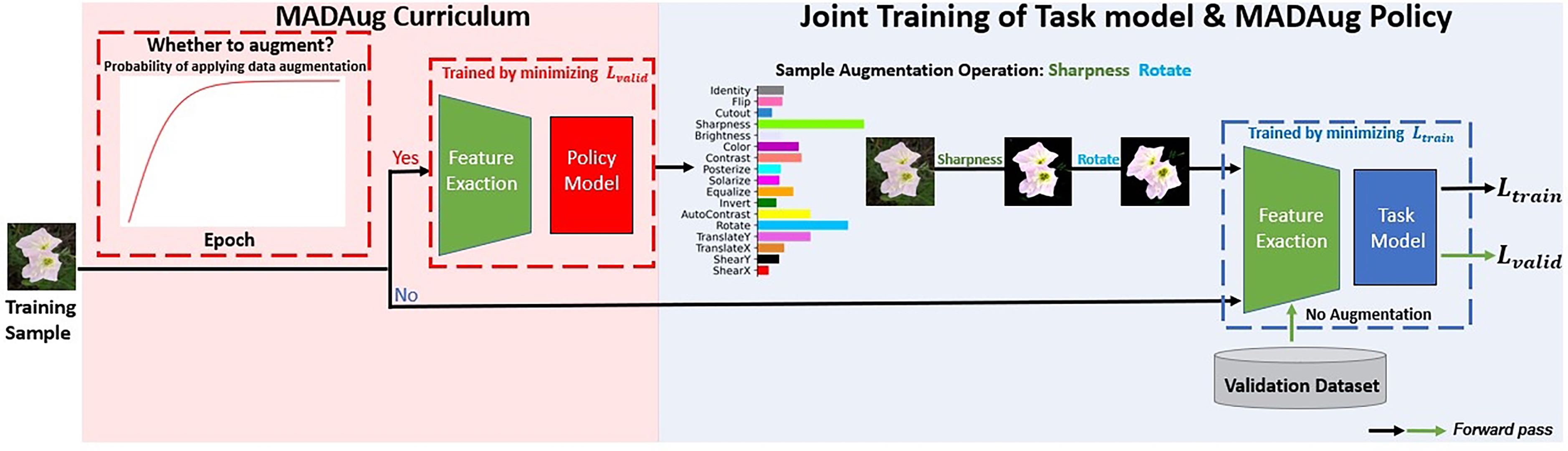}
    \caption{\textbf{\method} applies a~\trick to gradually introduce more data augmentations to the task model training and uses a policy network to choose augmentations for each training sample. \method trains the policy to minimize the validation loss of the task model, so the augmentations are model-adaptive and optimized for different training stages.  
    % In the early training stage, we take the original images as input. As the training progress, the task network gradually learns more augmented images. The policy network is trained by the performance of the task network on the validation dataset. 
    }
    \label{fig:basemethod}
\end{figure*}

First, applying data augmentation does not always bring improvement over the whole course of training. 
% it might not be a wise choice to constantly apply data augmentation during model training.
For example, we observed that a model tends to learn faster during earlier training stages without using data augmentation. 
% a model trained without data augmentation performs better than one trained with augmentation. 
We hypothesize that models at the early stage of training even have no capability to recognize the original images so excessively augmented images are not conducive to the convergence of the models. 
%As the training processes proceeds, increasing data augmentation can be chosen in order to learn more robust features.
Motivated by this observation, we first design a strategy called~\trick to progressively introduce more augmented data to the training.
% To achieve this goal,
In particular, we gradually increase the probability of applying data augmentation to each sample by following the \textit{Tanh} function (see Figure~\ref{fig:basemethod}), so the model can be quickly improved in earlier stages without distractions from augmentations while reaching a better performance in later stages through learning from augmented data. 
% we leverage the \textit{Tanh} function to increase the probability of applying data augmentation on specific samples with the training process (see Figure~\ref{fig:basemethod}).
% This strategy reaches high levels
% of performance on a model at the whole training stage.

%data transformations may only be applicable to certain domains, and heuristically selected transformations, such as transplanting transformations that are effective in one domain into another, may have the opposite effect~\cite{balestriero2022effects}. For example, the digit "0" is not sensitive to horizontal flipping with any degree, but the digit "6" becomes a digit "9" by further rotating it by 90 degrees.
%Other key consideration for applying data augmentation is that how to learn an augmentation policy for each image at different training processes, since redundant or overly aggressive augmentation can decease specific classes accuracy and slow down the training~\cite{balestriero2022effects}.
Secondly, a fixed augmentation policy is not optimal for learning every sample or different training stages. Although the~\trick gradually increases the augmented data as the model improves, it does not determine which augmentations applied to each sample can bring the most improvement to the model training.
Intuitively, the model can learn more from diverse data augmentations. 
% According to the above-mentioned study, we discover that the models should receive diverse data augmentations at various training stages.
Moreover, the difficulty of augmented data also has a great impact on the training and it depends on both the augmentations and the sample they are applied to. 
For example, ``simple" augmentation is preferred in the early stages to accelerate model convergence but more challenging augmented data provide additional information for learning more robust features for better generalization in the later stage.
One plausible strategy is leveraging expert knowledge and advice to adjust the augmentation operation and their strengths~\cite{mounsaveng2021learning,zhou2021metaaugment,shu2019meta,hataya2022meta}.
% A simple way to define adaptive data augmentation policies is to use expert knowledge to adjust the augmentation operations and
% their magnitude parameters ~\cite{mounsaveng2021learning,zhou2021metaaugment,shu2019meta,hataya2022meta}.
%However, as for a single dataset, an expert should be established to evaluate the availability of policy, which is not always possible due to cost
%constraints or limited expert knowledge.
%To  mitigate this challenge, regard the model performance on validation data as an expert to find the optimal data augmentation policy is a good choice~\cite{mounsaveng2021learning,zhou2021metaaugment,shu2019meta,hataya2022meta}. 
In this paper, instead of relying on human experts, we 
% In this paper, we 
regard the evaluation of the current model on a validation set as an expert to guide the optimization of augmentation policies applied to each sample in different training stages.
As illustrated in Figure~\ref{fig:basemethod}, we utilize a policy network to produce the augmentations for each sample (\ie, data-adaptive) used to train the task model, while the training objective of the policy network is to minimize the validation loss of the task model (\ie, model-adaptive). 
% Because the adjustment of policies is based on the current task model performance, we can call our policies as model-adaptive policies.
% The task model is optimized by minimizing the training loss, while the goal of the learned policies
% is to improve the performance of the task model on a validation set. 
This is a challenging bi-level optimization~\cite{colson2007overview}.
To address it, we train the task model on adaptive augmentations of training data and update the policy network to minimize the validation loss in an online manner. 
Thereby, the policy network is dynamically adapted to different training stages of the task model and generates customized augmentations for each sample. 
This results in a curriculum of data augmentations optimized for improving the generalization performance of the task model. 
% along with the training stage of a task model, rather than fixed throughout the whole training process. 
%Due to adopting the validation accuracy as expert knowledge, the computing cost and time
%overhead are extremely reduced. 
%Then, in each iteration, the augmentation policies can be learned and adjusted according to the current training phase of a task model.

Our main contributions can be summarized as follows:
\begin{enumerate}

\item [\textbf{(a)}] A~\textbf{\trick} gradually introducing more data augmentation to the training process.
% We find that using data augmentation at the initial training stages of model training does not improve the models' performance. 
% We introduce \trick to increase the probability of data augmentation with the training process.

\item [\textbf{(b)}] \textbf{\method} that trains a data augmentation policy network on the fly with the task model training. The policy automatically selects augmentations for each training sample and for different training stages.  

% We propose \method to learn the  model-adaptive data
% augmentation policies that dynamically adjust the augmentation policies according to the performance of the current task models.
\item [\textbf{(c)}] Experiments on CIFAR-10/100, SVHN, and ImageNet demonstrate that \method
consistently brings greater improvement to task models than existing data augmentation methods in terms of test-set performance. \looseness-1

\item [\textbf{(d)}] The augmentation policy network learned by \method on one dataset is \textbf{transferable to unseen datasets and downstream tasks}, producing better models than other baselines.
% such as Oxford Flowers, Oxford-IIT Pets, FGVC Aircraft, and Stanford Cars, when compared to other baselines.
\end{enumerate}

%% file: 02-related.tex
\section{Related Work}
Random crop and horizontal flip operations are commonly employed as standard data augmentation techniques for images in deep learning.
Recently, there are significant advancements in advanced data augmentation techniques that have significantly increased the accuracy of image recognition tasks~\cite{zhong2020random,yun2019cutmix,zhang2017mixup,tokozume2018between,devries2017improved,hendrycks2019augmix,hernandez2018data}.
However, data augmentations may only be applicable to certain domains, and heuristically selected transformations, such as transplanting transformations that are effective in one domain into another, could have the opposite effect~\cite{balestriero2022effects}. 
Thus, the exploration of optimal data augmentation policies necessitates specialized domain knowledge.

AutoAugment~\cite{cubuk2018autoaugment} adopts reinforcement learning to automatically find an available augmentation policy.
However, AutoAugment requires thousands of GPU hours
to find the policies on a reduced setting and limits randomness on the augmentation policies.
To tackle these challenges, searching the optimal data augmentation strategies has become a prominent research topic and many methods have been proposed ~\cite{zhong2020random,suzuki2022teachaugment,hataya2020faster,lim2019fast,hataya2022meta,li2020dada,lin2019online,li2020pointaugment,zhang2019adversarial,ho2019population,zheng2022deep,cheung2021AdaAug}.

These methods can be broadly classified into two distinct categories: fixed augmentation policies and online augmentation policies.
The first category of methods~\cite{hataya2020faster,li2020dada,lim2019fast,zhou2021metaaugment,cubuk2018autoaugment,zheng2022deep,cheung2021AdaAug} employs subsets of the training data and/or smaller models to efficiently discover fixed augmentation policies.
However, the limited randomness in these policies makes it challenging to generate suitable samples for various stages of training.
Thus, the fixed augmentation policies are suboptimal. 
The second category of methods~\cite{cubuk2020randaugment,suzuki2022teachaugment,lingchen2020uniformaugment,li2020pointaugment,muller2021trivialaugment,zhang2019adversarial,lin2019online}
focuses on directly finding dynamic augmentation policies on the task model. 
This strategy is increasingly recognized as the primary choice for data augmentation search.

RandAugment~\cite{cubuk2020randaugment} and TrivialAugment~\cite{muller2021trivialaugment} are typically the second type of methods for finding online augmentations.  
They randomly select the augmentation parameters without relying on any external knowledge or prior information. 
Other methods, such as Adversarial AutoAugment~\cite{zhang2019adversarial}, generate the adversarial augmentations by maximizing the training loss. 
However, the inherent instability of adversarial augmentations, without appropriate constraints, poses a risk of distorting the intrinsic meanings of images.
To avoid this collapse, TeachAugment~\cite{suzuki2022teachaugment} utilizes the “teacher knowledge” to effectively restrict adversarial augmentations.
However, Adversarial AutoAugment~\cite{zhang2019adversarial} and TeachAugment~\cite{suzuki2022teachaugment} both offer ``hard" augmentations rather than ``adoptive" augmentations, which are not effective to enhance the model generalization at the early training stage, because models at the early training even do not recognize the primitive images.
``Hard" augmentations are reluctant to converge the model.
Thus, in our paper,  we gradually apply the data augmentations for samples and track the model performance on the validation set to adjust the policies through the original bi-level optimization during the model training.
%\vspace{-1.0em}

%% file: 03-method.tex
\section{Method}

In this section, we first propose~\trick which progressively introduces more augmented samples as the training epoch increases. 
We then introduce the policy network that generates model-adaptive data augmentations and study how to train it through bi-level optimization with the task model.%\vspace{-0.5em}

\subsection{When to Augment: Monotonic Curriculum}
\label{sec;track}

\begin{figure}[h]
    \centering
\includegraphics[width=1.0\linewidth]{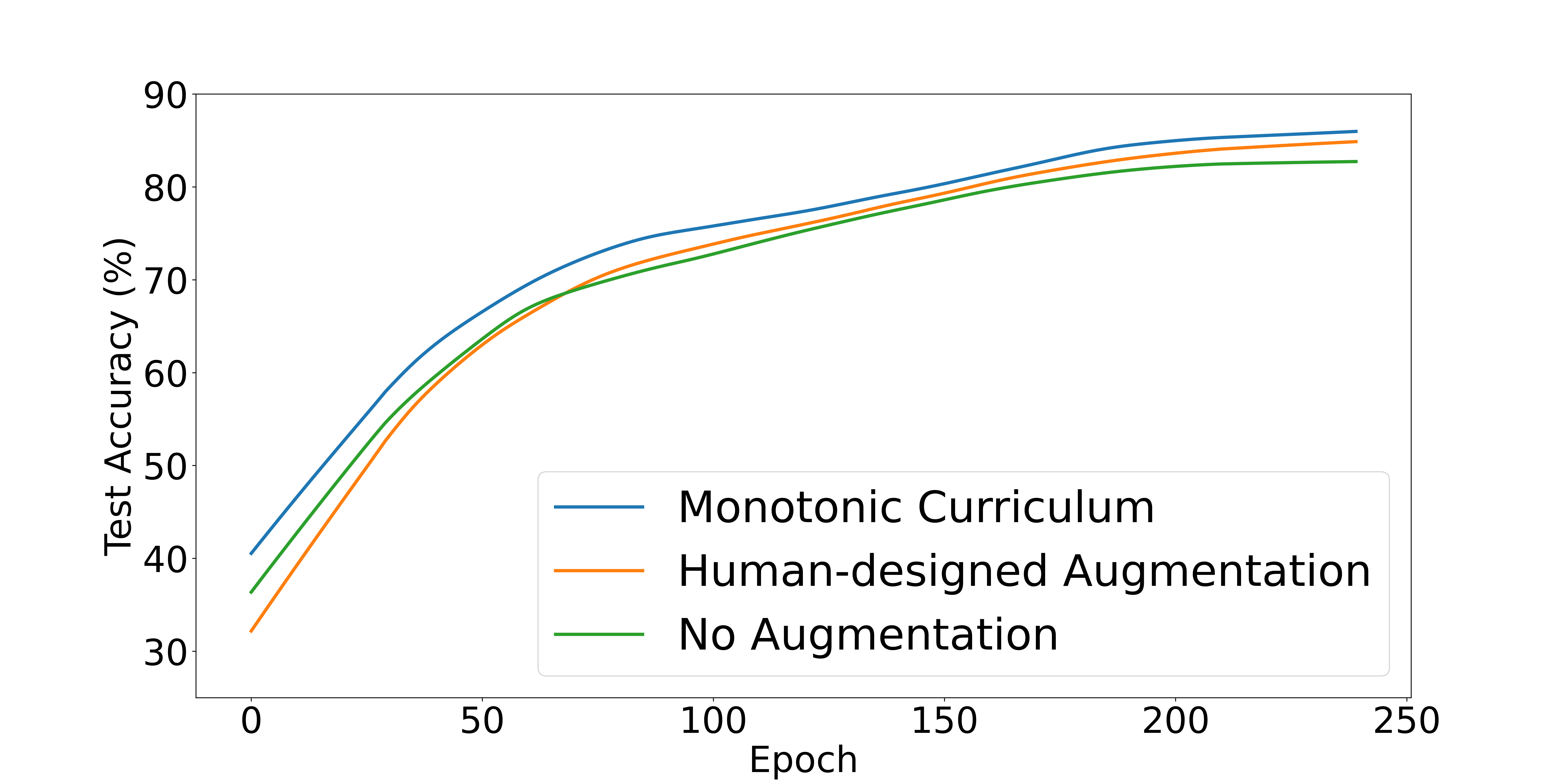}
    \caption{Test accuracy on Reduced CIFAR-10. \textbf{No Augmentation} does not apply any augmentations. \textbf{Human-designed Augmentation} always applies human pre-defined augmentations. \textbf{Monotonic Curriculum} gradually increases the probability of applying human-designed augmentations.}
    \label{fig: accuracy}
\end{figure} 

Previous studies~\cite{cubuk2018autoaugment,li2020dada,lim2019fast,hataya2020faster} have adopted the data augmentations for the whole model training process. 
However, at the early stage of model training, the model doesn't even recognize the original images.
In this case, is data augmentation effective?
In Figure~\ref{fig: accuracy}, the test accuracy of a model
trained on the Reduced CIFAR-10 dataset drops in the first $\sim 70$ epochs if applying human-designed data augmentations. 
To address this problem, at the beginning of model training, we only apply augmentations to a randomly sampled subset of training images while keeping the rest as original. In the later training stages, we apply a monotonic curriculum that gradually increases the proportion of images to be augmented or the probability of applying augmentation. 
Specifically, the proportion/probability $p(t)$ increases with the number of epochs by following a schedule defined by $\tanh$, i.e.,
\begin{equation}
p(t) = \tanh(t/\tau)
\label{eq:tanh}
\end{equation}
where $t$ is the current training epoch number and $\tau$ is a manually adjustable hyperparameter that controls the change of proportion. 
Therefore, the early-stage model is mainly trained on the original images without augmentations, which helps the premature model converge quickly.
However, as training proceeds, the model fully learned the original images and its training can benefit more from the augmented images. 
To validate the efficiency of our strategy,
compare with the images without augmentation policy or with the fixed human-design augmentation policies, our method can effectively boost model performance during various training stages (see Figure~\ref{fig: accuracy}).

\subsection{What Augmentations to Apply: Model-Adaptive Data Augmentation}

Instead of constantly applying the same data augmentation policies to all samples over the whole training process, adjusting the policy for each sample and model in different training stages can provide better guidance to the task model and thus accelerate its training towards better validation accuracy. 

% model-adaptive and data-dependent augmentation policies can further accelerate the validation accuracy of the task model.

Following  AdaAug~\cite{cheung2021AdaAug},
%we formulate its augmentation policy with several augmentation operations for each image (\eg, flip).
we assign an augmentation probability $p$ and magnitude $\lambda$ to each sample.
The augmentation probability vector $p$ contains the possibility $p_i$ of applying each augmentation-$i$, \ie, $\sum_{i=1}^n p_i =1$,  where there are $n$ possible augmentation operations. The augmentation magnitude vector $\lambda$ contains the associated augmentation strengths such that $\lambda_i \in [0,1]$.
% The magnitude parameter $\lambda$ specifies the degree of the transformation.
In the training process, for every training image $x$, we draw $k$ operations without replacement according to $p$ and build an augmentation policy based on them and their magnitude in $\lambda$. 
In particular, each sampled augmentation operator-$j$ is applied to the image $x$ with magnitude $\lambda_{j}$, resulting in an augmented image $\Gamma^{j}(x)\triangleq\tau_j(x;\lambda_{j})$. By applying the $k$ sampled augmentations, the final augmented image $\gamma(x)$ can be written as:
\begin{equation}
	\begin{split}
	&\Gamma^{t}(x) = \tau_j(x; \lambda_{j}), \quad  j  \sim p\\
	 &\gamma(x) = \Gamma^{k}\circ \cdots \circ \Gamma^{1} (x),
	\end{split}
\end{equation}
where $\circ$ is the compositional operator. 
% and $\Gamma^{t}(x)$ is a single transformation applied on an image $x$ with $t$-th augmentation operation in a policy.  
%Our goal is to provide an adaptive augmentation policy for each image to enhance the performance of model on the validation set at the different stage of training process (see Figure~\ref{fig:basemethod}).

An arbitrary augmentation policy is not guaranteed to improve the performance of a task model but a brute-force search is not practically feasible. Hence, we optimize a policy model producing the optimal augmentation probability vector $p$ and magnitude vector $\lambda$ for each image at different training stages.
For image $x$, we define $f(x;w)$ as the task model with parameter $w$ and $g_{w}(x)$ as the intermediate-layer representation of image $x$ extracted from the task model $f(x;w)$. 
The policy model $p(\cdot;\theta)$ with parameters $\theta$ takes the extracted feature $g_{w}(x)$ as input
and outputs the probability vector $p$ and magnitude vector $\lambda$ for the image $x$.
The parameter $w$ of the task model is optimized by minimizing the following training loss on the training set $\mathcal{D}^{tr} = \left\{x_i,y_i\right\}_{i=1}^{N^{tr}}$:
\begin{equation}
\centering
w =\mathop{\arg\min}\limits_{w} \mathcal{L}^{tr}(w;\theta) =\frac{1}{N^{tr}}\sum_{i=1}^{N^{tr}} \mathcal{L}_{CE}(f(\gamma(x_{i});w) ,y_{i}),
\label{eq:trainloss}
\end{equation}
 where the augmented training image $\gamma(x_{i})$ is generated by the policy network $p(g_{w}(x_{i}) ;\theta)$ and $\mathcal{L}_{CE}( \cdot ,\cdot)$ is the cross-entropy loss.
 The policy model is to produce augmentation policies applied to the training of the task model and its optimization objective is to minimize the trained task model's loss on a validation set, i.e., $\mathcal{D}^{val} = \left\{x_i^{val},y_i^{val}\right\}_{i=1}^{N^{val}}$.  %In the inner loop, the parameters $w$ of the task network are optimized on the training data $\mathcal{D}^{tr}$ according to the current policy network.  
 The above problem can be formulated as the bi-level optimization~\cite{colson2007overview} below:
\begin{equation}
	\begin{split}
	&\min_{\theta} \quad  \mathcal{L}^{val}(w^{\ast}(\theta)) = \frac{1}{N^{val}} \sum_{i=1}^{N^{val}}\mathcal{L}^{val}_{i}(w^{\ast}(\theta)) \\
	 &s.t. \quad w^{\ast}(\theta)= \arg\min_{w} \mathcal{L}^{tr}(w;\theta)
	\end{split}
\label{eq:optimaiza}
\end{equation}
where $\mathcal{L}^{val}_{i}(w^{\ast}(\theta))= \mathcal{L}_{CE}(f(x_i^{val};w^{\ast}(\theta)) ,y)$.
% This is a typical 
% In the bilevel optimization, it is usually challenging to simultaneously optimize the policy model's parameters $\theta$ and the task model's parameters $w$. 
Bi-level optimization is challenging because the lower-level optimization (i.e., the optimization of $w$) does not have a closed-form solution that can be substituted into the higher-level optimization (i.e., the optimization of $\theta$). 
Recent work~\cite{ren2018learning,shu2019meta,zhou2021metaaugment,hataya2022meta,mounsaveng2021learning} address this problem (\ref{eq:optimaiza}) by alternating minimization. 
In this paper, we employ the same strategy as~\cite{shu2019meta,zhou2021metaaugment,antoniou2018train}.

\subsection{Joint Training of Task model \& MADAug Policy}

To address the bi-level optimization, we alternately update $\theta$ and $w$ by first optimizing the policy network $\theta$ for a task model $\hat w$ achieved by one-step training and then update $w$ using the augmentations produced by the new policy network $\theta$.

% The policy and task models are trained alternately.
We split the original training set into two disjoint sets, \ie, a training set and a validation set.
Each iteration trains the model on a mini-batch of $n^{tr}$ images $\mathcal{D}^{tr}_{m_i} = \left\{x_i,y_i\right\}_{i=1}^{n^{tr}}$ drawn from the training set. 
Let $\mathcal{L}^{tr}(w_{t};\theta_{t})=\mathcal{L}_{CE}(f(\gamma(x_{i});w_{t}),y_i)$ denote the lower-level objective for optimizing $w_t$. We apply one-step gradient descent on $w_t$ to achieve a closed-form surrogate of the lower-level problem solution, i.e.,
\begin{equation}
\hat{w_{t}} = w_{t}-\alpha \frac{1}{n^{tr}}\sum_{i=1}^{n^{tr}} \nabla_{w}\mathcal{L}^{tr}(w_{t};\theta_{t})
\label{eq;train}
\end{equation}
where $\alpha$ is a learning rate. 
However, we cannot use back-propagation to optimize $\theta_t$ for the high-level optimization because the sampling process of the $k$ augmentation operations in $\gamma(x_i)$ is non-differentiable. Hence, back-propagation cannot compute the partial derivative w.r.t. the augmentation probability $p$ and magnitude $\lambda$. 
To address this problem, we relax the non-differentiable $\gamma(x_i)$ to be a differentiable operator. Since the augmentation policy in most previous work~\cite{cubuk2018autoaugment,ho2019population} only consists of two operations, for $k=2$, $\gamma(x_i)$ can be relaxed as
% The differentiable variable can be changed as:

%\begin{small}
%\begin{equation}
%\begin{split}
%&\gamma(x_i) \approx \sum_{j_{k}=1}^n %\sum_{j_{k-1}=1}^n\cdots\sum_{j_{1}=1}^n p_{ij_{k}} \cdot p_{ij_{k-1}}\cdots p_{ij_{1}}  \Gamma_{ij_{k}}( \Gamma_{ij_{k-1}}(\cdots( \Gamma_{ij_{1}}(x_i))))\\ 
%&\quad j_{k}\neq j_{k-1}\neq\cdots \neq j_{1}
%\end{split}
%\label{eq;aug}
%\end{equation}
%\end{small}

\begin{equation}
\gamma(x_i) \approx \sum_{j_{1}=1}^n \sum_{j_{2}=1}^n p_{ij_{1}} \cdot p_{ij_{2}} \Gamma_{ij_{2}}^{2}( \Gamma_{ij_{1}}^{1}(x_i))))
\quad j_{1}\neq j_{2}
\label{eq;aug}
\end{equation}
where $\Gamma_{ij_{k}}^{t}(x_i)=\tau_{j_{k}}(x_{i}; \lambda_{j_{k}})$ applies augmentation-$j_{k}$ (with magnitude $\lambda_{j_{k}}$) to $x_i$ in the $t$-th augmentation operation.
% $\lambda_{j_{k}}$ is the magnitude of augmentation $j_{k}$, which is also generated by the policy model.
The relaxed $\gamma(x_i)$ is differentiable by combining different augmentations according to weights as their probabilities, so we can estimate the partial derivatives w.r.t. $p$ via back-propagation through Eq.~\ref{eq;aug}. In our approach, the forward pass still uses the sampling-based $\gamma(x_i)$, whereas the backward pass uses its differentiable relaxation in Eq.~\ref{eq;aug}. 

For back-propagation through the augmentation magnitude vector $\lambda$, we apply the straight-through gradient estimator~\cite{bengio2013estimating,van2017neural} because the magnitudes of some operations such as ``Posterize'' and ``Solarize'' are discrete variables that only have finite choices.
%Specifically, the magnitudes' gradient is estimated in relation to  each input pixel value $x_{h,w}$ from its augmented image $r(x)$, \ie $\frac{\partial \gamma(x_{h,w}) }{\partial \lambda_{j}}=1$. 
In previous approaches~\cite{cheung2021AdaAug,hataya2020faster}, the loss's gradient w.r.t. $\lambda_m$ is estimated by applying the chain-rule to each pixel value $\gamma(x_{h,w})$ in the augmented image $\gamma(x)$, \ie $\frac{\partial \gamma(x_{h,w}) }{\partial \lambda_{j}}=1$. Hence, the gradient of loss $\mathcal{L}$ w.r.t. $\lambda_m$ can be computed as:
\begin{equation}
\frac{\partial \mathcal{L} }{\partial \lambda_{m}}=\sum_{h,w}\frac{\partial \mathcal{L} }{\partial \gamma(x_{h,w})}\frac{\partial \gamma(x_{h,w}) }{\partial \lambda_{m}}=\sum_{h,w}\frac{\partial \mathcal{L} }{\partial \gamma(x_{h,w})}
\label{eq;lambda}
\end{equation}

%We optimize augmentation magnitudes $\lambda$ using the straight-through gradient estimator technique~\cite{bengio2013estimating,van2017neural}. This is because some operations, such as Posterize and Solarize, involve the discretization of magnitudes. In this approach, the gradient of the magnitudes is estimated with respect to each input pixel value $x_{h,w}$ from its augmented image $r(x)$. Specifically, the partial derivative of $\gamma(x_{h,w})$ with respect to $\lambda_{j}$ is estimated as $\frac{\partial \gamma(x_{h,w}) }{\partial \lambda_{j}}=1$. 
%Thus, the gradient of magnitude $\lambda_m$ can be calculated by:
%\begin{equation}
%\frac{\partial \mathcal{L} }{\partial \lambda_{m}}=\sum_{h,w}\frac{\partial \mathcal{L} }{\partial \gamma(x_{h,w})}\frac{\partial \gamma(x_{h,w}) }{\partial \lambda_{m}}=\sum_{h,w}\frac{\partial \mathcal{L} }{\partial \gamma(x_{h,w})}
%\label{eq;lambda}
%\end{equation}

Then, the policy network parameters $\theta_t$ can be updated by minimizing the validation loss computed by the current meta task model $\hat{w_{t}}$ on a mini-batch of validation
set $\mathcal{D}^{val} = \left\{x_i^{val},y_i^{val}\right\}_{i=1}^{n^{val}}$ with batch size $n^{val}$. 
Therefore, the
outer loop updates of $\theta_t$ is formulated by:
\begin{equation}
\theta_{t+1} = \theta_{t}-\beta \frac{1}{n^{val}}\sum_{i=1}^{n^{val}}\nabla_{\theta}\mathcal{L}^{val}_{i}(\hat{w_{t}}(\theta_{t}))
\label{eq:val}
\end{equation}
where $\beta$ is a learning rate. 
The third step is to update the parameter $w_{t}$  based on the  parameter $\theta_{t+1}$ of the policy model in the outer loop of iteration $t + 1$:
\begin{equation}
w_{t+1} = w_{t}-\alpha \frac{1}{n^{tr}}\sum_{i=1}^{n^{tr}}\nabla_{w}\mathcal{L}^{tr}(w_{t};\theta_{t+1})
\label{eq:entire}
\end{equation}
With these updating rules, the policy and task
networks can be alternatively trained.
Our proposed algorithm is summarized in Algorithm~\ref{alg:Framwork}.

\begin{algorithm}[h]  
  \caption{Model-Adaptive Data Augmentation}
  \label{alg:Framwork}  
  \begin{algorithmic}[1]  
   \Require 
      Training set $\mathcal{D}_{train}=\left\{x_i,y_i\right\}_{i\in [N_{train}]}$; 
      Validation set $\mathcal{D}_{vaild}=\left\{x_i,y_i\right\}_{i\in [N_{valid}]}$;
       Batch sizes $ n^{tr}, n^{val}$;
       Learning rate $\alpha, \beta$;
       Iteration number $T$;
   \Ensure Task model $w_{T}$; policy network $\theta_{T}$
   \State Initialize $w_{0}$, $\theta_{0}$
    \For{$t= 0$ to $T$}
        \State Sample a training set mini-batch $d_{train}\in \mathcal{D}_{train}$.
        %\State Augment the train sample based on the current  policy network with $K$ augmentations
        \State Draw Augment$\sim P(t)$ in Eq.~\ref{eq:tanh}.
        \If{Augment}
        \State Apply policy network $\theta_{t}$ to achieve augmentations $\gamma(x)$ for each sample $x\in d_{train}$. 
        \EndIf
        \State Update $\hat{w_{t}}$ on the augmented $d_{train}$ (Eq.\ref{eq;train}).
        \State Sample a validation set mini-batch $d_{valid}\in \mathcal{D}_{valid}$.
        \State Update policy network $\theta_{t+1}$ on $d_{valid}$ (Eq.~\ref{eq:val}).
        \State Apply policy network $\theta_{t+1}$ to achieve new augmentations $\gamma(x)$ for each sample $x\in d_{train}$. 
        \State Update task model $w_{t+1}$ on the newly augmented $d_{train}$ (Eq.~\ref{eq:entire}).
        
    \EndFor 
  \end{algorithmic}  
\end{algorithm}

%% file: 04-exp.tex
\section{Experiments}
\begin{table*}[t]
\large
\centering
\caption{\textbf{Test error (\%, average of 5 random trials) on CIFAR-10, CIFAR-100,  SVHN and ImageNet.} Lower value is better. ``Simple'' applies regular random crop, random horizontal flip, and Cutout. All other methods apply ``Simple'' on top of their proposed augmentations. 
% Cutout is also applied on top of the baselines, just like PBA/AutoAugment. We report the mean final test error of 5 random model initializations. We used the models: Wide-ResNet-40-2, Wide-ResNet-28-10, Shake-Shake (26 2x96d),  PyramidNet
%with ShakeDrop and ResNet-50. 
We report the accuracy of our re-implemented AdaAug$\dagger$, while other baselines' results are adapted from Zheng \etal~\cite{zheng2022deep}, Cheung \etal~\cite{cheung2021AdaAug}, Tang \etal~\cite{tang2020onlineaugment}, and Suzuki \etal~\cite{suzuki2022teachaugment}. The best performance is highlighted in \textbf{Bold}.}
%\begin{adjustbox}{max width=1.0\textwidth}
\resizebox{1.0\textwidth}{!}{
\resizebox{3.0\linewidth}{!}{
\begin{tabular}{ll|ccccccccccccccc}
\hline
Dataset  &Backbone& Simple  &AA & PBA &Fast AA&DADA&Faster AA &RA&TA&DeepAA&Teach&OnlineAug&AdaAug&AdaAug$\dagger$&\method\\\cline{1-16}
\multirow{2}*{Reduced CIFAR-10}&Wide-ResNet-28-10& 18.9&14.1 &12.8 &14.6 &15.6 &- &15.1&-&-&-&14.3& 13.6 &15.0&\textbf{12.5}\\
~&Shake-Shake (26 2x96d)  &17.1 &10.1&10.6 &-&- &-&-&-&- &-&-&10.9 &11.8&\textbf{10.0}\\
\hline
\multirow{4}*{CIFAR-10}& Wide-ResNet-40-2&5.3 &3.7 &3.9& 3.6 &3.6 &3.7 &4.1&-&-&-&-&3.6&-&\textbf{3.3}\\
~&Wide-ResNet-28-10& 3.9 &2.6 &2.6 &2.7 &2.7 &2.6 &2.7&2.5&2.4&2.5&2.4 &2.6&-&\textbf{2.1}\\
~&Shake-Shake (26 2x96d)&2.9 &2.0 &2.0 &2.0 &2.0 &2.0 &2.0&1.9&1.9&2.0&- &-&-&\textbf{1.8}\\
~&PyramidNet (ShakeDrop)&2.7 &1.5 &1.5 &1.8 &1.7 &- &1.5&-&-&1.5&- &-&-&\textbf{1.4}\\
\hline
\multirow{4}*{CIFAR-100}& Wide-ResNet-40-2&26.0 &20.6 &22.3 &20.7 &20.9 &21.4&-&19.4&- &-&- &19.8&-&\textbf{19.3}\\
~&Wide-ResNet-28-10& 18.8 &17.1 &16.7 &17.3 &17.5 &17.3 &16.7&16.5&16.1&16.8&16.6 &17.1&-&\textbf{16.1}\\
~&Shake-Shake (26 2x96d)&17.1 &14.3 &15.3 &14.9 &15.3 &15.6  &- &-&14.8&14.5&- &-&-&\textbf{14.1}\\
~&PyramidNet (ShakeDrop)&14.0 &10.7  &10.9 &11.9 &11.2&- &- &-&-&11.8&- &-&-&\textbf{10.5}\\
\hline
\multirow{2}*{Reduced SVHN }&Wide-ResNet-28-10&13.2 &8.2 & 7.8 &8.1 &7.6 &- &9.4&-&-&-&\textbf{6.7} &8.2 &9.1&8.4\\
~&Shake-Shake (26 2x96d)&13.3 &\textbf{5.9}&6.5 &- &- &- &- &-&-&-&- &6.4&6.9&6.3\\
\hline
\multirow{2}*{SVHN} &Wide-ResNet-28-10&1.5 &1.1 &1.2& 1.1&1.2& 1.2&1.0&-&-&-&-&-&-&\textbf{1.0}\\
~&Shake-Shake (26 2x96d)&1.4 &1.1 &1.1 & 1.1&1.1 &- &- &-&-&-&- &-&-&\textbf{1.0}\\
\hline
ImageNet &ResNet-50 &23.7 &22.4 &-& 22.4&22.5& 23.5&22.4&22.1&21.7&22.2&22.5  &22.8&-&\textbf{21.5}\\
\hline
\label{tab:benchmark}
\end{tabular}}}
%\end{adjustbox}
\end{table*}

% \subsection{Summary of experiments}
In this section, following AutoAugment~\cite{cubuk2018autoaugment}, we examine the performance of \method on two experiments: \method-direct and \method-transfer.
In the first experiment, we directly explore the performance of  the \method 
on the benchmark datasets: CIFAR-10~\cite{krizhevsky2009learning}, CIFAR-100~\cite{krizhevsky2009learning}, SVHN~\cite{netzer2011reading}, and ImageNet~\cite{deng2009imagenet}.
For CIFAR-10, CIFAR-100, and SVHN, we equally select 1,000 images from the dataset as the validation set to train the policy model.
Plus, for SVHN, we apply both the training images and additional “extra” training images as the training set. 
For ImageNet, the validation set consists of 1,200 examples from a randomly selected 120 classes.
We compare the average test set error of our method with previous state-of-the-art methods, AutoAugment (AA)~\cite{cubuk2018autoaugment}, Population Based Augmentation (PBA)~\cite{ho2019population}, Fast AutoAugment (Fast AA)~\cite{lim2019fast}, DADA~\cite{li2020dada}, Faster AutoAugment (Fasrer AA)~\cite{hataya2020faster},
RandAugment (RA)~\cite{cubuk2020randaugment}, 
TrivialAugmen (TA)~\cite{muller2021trivialaugment},
Deep AutoAugment (Deep AA)~\cite{zheng2022deep}, TeachAugment (Teach)~\cite{suzuki2022teachaugment}, OnlineAug~\cite{tang2020onlineaugment}, and AdaAug~\cite{cheung2021AdaAug}. 

Our experiment results  demonstrate that \method-direct considerably improves the accuracy of baselines and achieves state-of-the-art performance on these benchmark datasets. 
In the second experiment, we investigate the transferability of \method-learned policy network to unseen fine-grained datasets. To verify its effectiveness, we apply the augmentation policies learned by \method on the CIFAR-100 dataset to fine-grained classification datasets such as Oxford 102 Flowers~\cite{nilsback2008automated}, Oxford-IIIT Pets~\cite{em2017incorporating}, FGVC Aircraft~\cite{maji2013fine}, and Stanford Cars~\cite{krause2013collecting}. 
Our findings demonstrate the remarkable transferability of \method-learned policy, which significantly outperforms the robust baseline models on fine-grained classification datasets.

\subsection{Augmentation Operations}
We follow the augmentation actions taken by AutoAugment~\cite{cubuk2018autoaugment}. We adopt the 16 augmentation operations (ShearX, ShearY, TranslateX, TranslateY, Rotate, AutoContrast, Invert, Equalize, Solarize, Posterize, Contrast, Color, Brightness, Sharpness, and Cutout) that are previously suggested to build the augmentation policies.
Meanwhile, we add the Identity operation, which does not apply  augmentation on images. 
For the sample baseline, we employ random horizontal flip, color jittering, color normalization, and Cutout with a $16\times16$ patch size as basic augmentations. 
The found policies learned by \method and other baselines are applied on top of these basic augmentations.

\begin{table*}[t]
\centering
\caption{\textbf{Transferability of \method learned policy network.} Test set error (\%) of fine-tuning a pertrained ResNet-50 using the augmentations produced by the policy network on downstream tasks. 
% We also apply random crop, horizontal flip operations and Cutout on the top of different augmentations. 
Baseline results are adapted from Cheung \etal~\cite{cheung2021AdaAug}.\looseness-1}
\begin{adjustbox}{max width=1.0\textwidth}
\begin{tabular}{lcccccccc}
\hline
 Dataset & \# of classes &Train number & Simple  &AA  &Fast AA&RA&AdaAug&\method\\
 \hline
Oxford 102 Flowers&102&2,040&5.0 &6.1 &4.8 &3.9 &2.8 &\textbf{2.5}\\
Oxford-IIIT Pets&37&3,680&19.5 &18.8&23.0 &16.8 &16.1&\textbf{15.3}\\
FGVC Aircraft&100&6,667& 18.4 &16.6 &17.0 &17.4 &16.0&\textbf{15.4}\\
Stanford Cars&196&8,144&11.9 &9.2 &10.7 &10.3 &8.8&\textbf{8.3}\\
\hline
\label{tab:finegrained dataset}
\end{tabular}
\end{adjustbox}
\end{table*}

\subsection{Implementation Details}
In our experiments, the policy network of~\method refers to a fully-connected layer that takes the representations produced by the penultimate layer of the task model as its inputs and outputs $p$ and $\lambda$.
% In our experiments, we use the task model's penultimate layer the feature extraction network is the current task model. 
% Then, we implement the policy projection network as a linear layer, which inputs the extracted features from the current task model and outputs the operation parameters  $p$ and $\lambda$.
Following AdaAug~\cite{cheung2021AdaAug}, the update of policy projection network parameters uses the Adam optimizer with a learning rate of $0.001$.
For the CIFAR-10, CIFAR-100, and SVHN, we evaluate our method on four models: Wide-ResNet-
40-2~\cite{zagoruyko2016wide}, Wide-ResNet-28-10~\cite{zagoruyko2016wide},
Shake-Shake (26 2x96d)~\cite{gastaldi2017shake}, and PyramidNet with ShakeDrop~\cite{yamada2019shakedrop,han2017deep}.
We train all models using a batch size of 128 except for PyramidNet with Shake-Drop, which is trained with a batch size of 64.
We train the Wide-ResNet for 200 epochs and Shake-Shake/PyramidNet for 1,800 epochs.
For Wide-ResNet models trained on SVHN, we follow PBA~\cite{ho2019population} to use the step learning rate schedule~\cite{devries2017improved} and all other models  use a cosine learning rate scheduler with one annealing cycle~\cite{loshchilov2016sgdr}.
% In ImageNet experiments, we sample 1,200 training images from 120 randomly chosen classes as the validation set for training the policy network. 
% The hyperparameters of the policy model on ImageNet are the same as those set on CIFAR-10. 
To align our results with other baselines, we train the ResNet-50~\cite{he2016deep} from scratch on the full ImageNet using the hyperparameters in AutoAugment~\cite{cubuk2018autoaugment} on ImageNet.
For all models, we use gradient clipping with magnitude 5. 
We provide specific details about the learning rate and weight decay values on the supplementary materials.

\subsection{Main Results} 

Table~\ref{tab:benchmark} shows that the learned policies through bi-level optimization achieve the
best performance than the baselines for different models on the Reduced CIFAR-10, CIFAR-10, CIFAR 100, Reduced SVHN, SVHN, and ImageNet. 
The Reduced CIFAR-10(SVHN) dataset randomly selects 4,000(1,000) images for CIFAR-10(SVHN) as the training set and sets the remaining images as the validation set.
% We also evaluate \method on the SVHN dataset, as shown in Table~\ref{tab:benchmark}.
\method achieves state-of-the-art performance on this dataset.
On the Reduced SVHN dataset, compared to AdaAug, we achieve 0.7\% and 0.6\% improvement on Wide-ResNet-28-10 and Shake-Shake (26 2x96d), respectively. 
On ImageNet, compare with other baselines, our method performs the best on a large and complex dataset. 
Different from the prior work (AutoAugment, PBA, and Fast AutoAugment) which constructs the fixed augmentation policies for the enter dataset, our method can find the dynamic and
 model-adoptive policies for each image, which enhances the model's generalization.  We provide the average and variance of the experiment results in Section~\ref{app:var}.

\subsection{Transferability of \method-Learned Policy}

Following AdaAug~\cite{cheung2021AdaAug}, we apply the augmentation policies learned from the CIFAR-100 directly on the fine-grained datasets (\method-direct). 
To evaluate the transferability of the policies found on CIFAR-100, we compare the test error rate with AutoAugment (AA)~\cite{cubuk2018autoaugment}, Fast AutoAugment (Fast AA)~\cite{lim2019fast}, RandAugment (RA)~\cite{cubuk2020randaugment}, and AdaAug~\cite{cheung2021AdaAug} using their published policies on CIFAR-100.
For all the fine-grained datasets, we compare the transfer results by training the ResNet-50 model~\cite{he2016deep} pretrained on ImageNet. Following the experiment setting of AdaAug, we use the cosine learning rate decay with one annealing cycle~\cite{loshchilov2016sgdr} and train the model for 100 epochs.
According to the validation performance, we adjust the learning rate for different fine-grained datasets. The weight decay is set as 1e-4 and the gradient clipping parameter is 5.

Table~\ref{tab:finegrained dataset} shows that our method outperforms the other baselines when training the
 pertrained ResNet-50 model on these fine-grained datasets.
Previous methods (AutoAugment~\cite{cubuk2018autoaugment}, Fast Augmentation~\cite{lim2019fast}, and RandAugment~\cite{cubuk2020randaugment})  apply the fixed augmentation policies. 
This strategy does not help the fine-grained sample to distinguish from each other, which makes the model easy to recognize their differences.
In contrast, AdaAug 
 and \method adapt the augmentation policies for the entire dataset to instance-level grade.
 Because our method gradually augments the images and provides more optimal augmentation policies
to unseen fine-grained images according to their relationship to the classes that have been learned on the CIFAR-100 dataset, the model achieves better performance than AdaAug, especially for the Pet dataset. The Pet dataset only contains the ``Cat" and ``Dog" images.
In Figure~\ref{fig: per-class accuracy}, we also show that \method can improve the model's ability to recognize ``Cat" and ``Dog" classes significantly on the Reduced CIFAR-10 dataset.

%% file: 05-discussion.tex
%\section{Discussion}
\begin{figure}[h]
    \centering
    \includegraphics[width=1.0\linewidth]{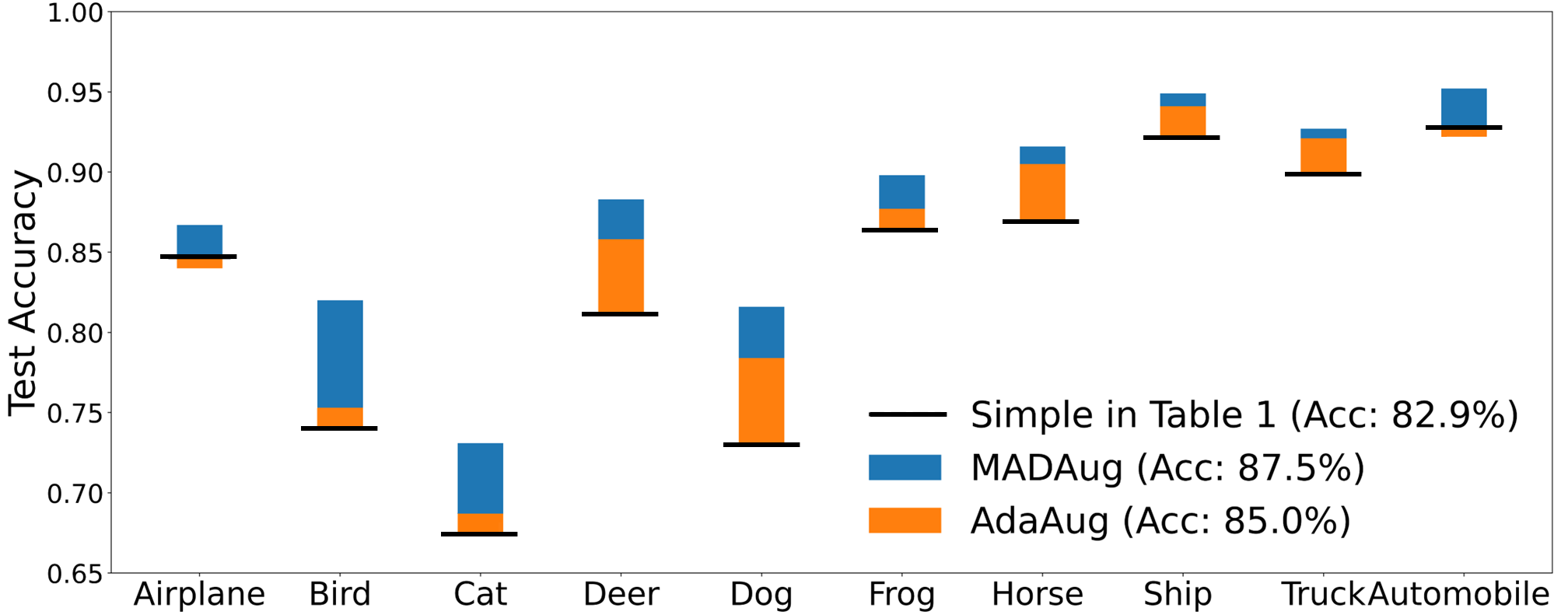}
    \caption{\textbf{Improvement that \method and AdaAug bring to different classes.} 
    % The black line is sample baseline accuracy only with basic augmentation.
    \method consistently improves the test accuracy over all classes and brings greater improvements to more difficult classes (fairness), \eg, ``Bird", ``Cat", ``Deer", and ``Dog". In contrast, AdaAug has a negative impact on ``Airplane" and ``Automobile".
    % considerably increases accuracy in the ``Bird", ``Cat", ``Deer", and ``Dog" classes. Data augmentation learned by AduAug has a negative impact  on the ``Airplane" and ``Automobile" classes.
    }
    \label{fig: per-class accuracy}
   
\end{figure} 

\subsection{Analysis of \method Augmentations}

\begin{figure*}[t]
    \centering
\includegraphics[width=0.92\linewidth]{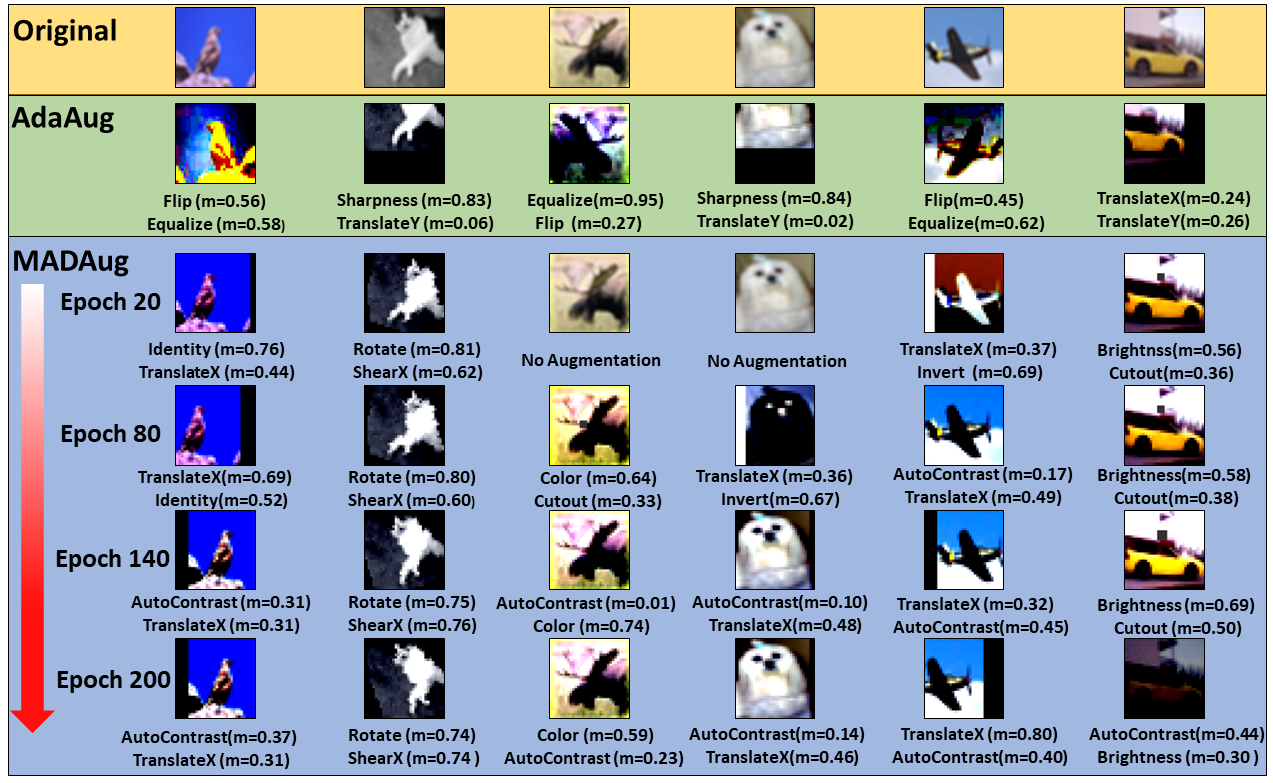}   
    \caption{\textbf{Augmentations of AdaAug and \method for different classes of images (operations and associated strengths).} AdaAug only produces specific augmentations for different images, while~\method adjusts the augmentations for each image to be adaptive to different training epochs. \method introduces less distortion than AdaAug.
    % Data augmentations learned by AdaAug and \method are applied to different class images on the Reduced CIFAR-10 dataset.  Each augmentation is provided with its operation and magnitude respectively.
    % \method produces a wide range of accurate augmented images without distorting their original meanings.
    }
    \label{fig: polciy vs}
\end{figure*}

 We compare the pre-class accuracy from \method with AdaAug on the Reduced CIFAR-10 dataset, which only has 4,000 training images.
Figure~\ref{fig: per-class accuracy} shows the pre-class accuracy of a model trained on \method is higher than that trained on AdaAug and basic baseline, especially in ``Bird", ``Deer", ``Cat" and ``Dog" classes.
Moreover, compared with the basic baseline, we can see that the augmentation policies trained by AdaAug play a negative impact on the ``Airplane" and ``Automobile" classes. 
%\textcolor{blue}{We think that the decrease in classification accuracy across different categories in the AdaAug method can be attributed to the long-tail distribution of semanticinformation caused by  inadequate data augmentation strategies~\cite{balestriero2022effects}.}

\begin{figure}[h]
    \centering
\includegraphics[width=1.0\linewidth]{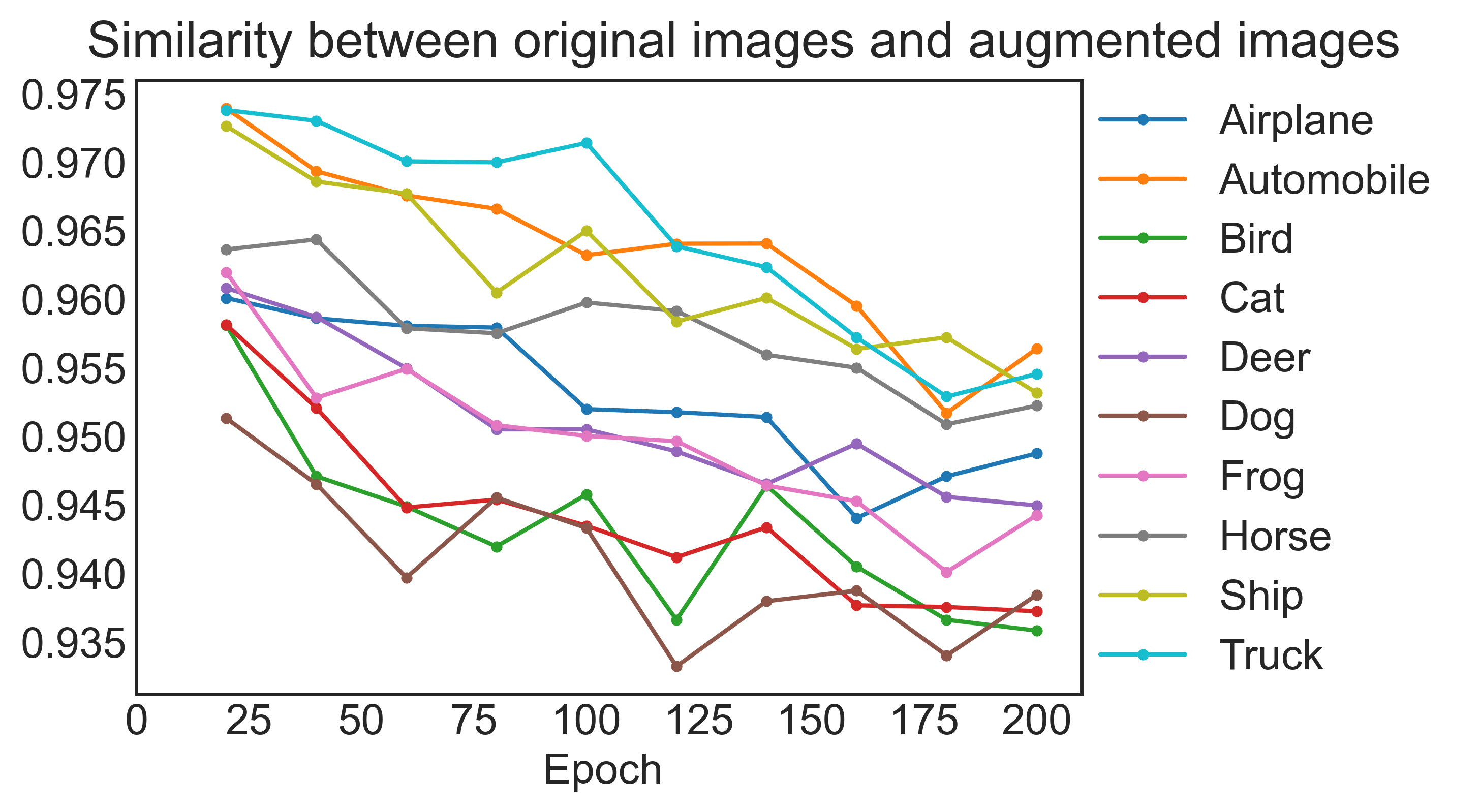}
    \caption{\textbf{Similarity between the original images and \method-augmented images at different training epochs.} \method starts from less perturbed images but generates more challenging augmentations in later training. }
    \label{fig: simility}

\end{figure}

In Figure~\ref{fig: polciy vs}, we display some augmented samples with AdaAug and \method, which are randomly selected from ``Bird", ``Cat", ``Deer", ``Dog", ``Airplane", and ``Automobile" classes. 
%Similar transformations may have very different effects on different images. 
For AdaAug, some augmented images in their classes have
lost semantic information caused by the translation, because augmentation operations like TranslateY with unreasonable magnitude collapse the main information of the image. 
For example, in the Figure~\ref{fig: polciy vs}, the selected ``Cat" augmented image loses its face and only leaves its legs. 
And the ``Dog" augmented image also discards a part of key information.
We think that these unreasonable data augmentation strategies lead to an  imbalance in the number of samples containing sufficient information about their true label across different categories, although the original dataset is balanced~\cite{balestriero2022effects}. This phenomenon leads to a reduction in the classification accuracy of some categories.
However, for our method, Figure~\ref{fig: simility}
shows augmentation policies generated by  \method can produce more ``hard" samples for the model with the training process.
This strategy would improve the model generalization because at the early phase, ``simple" samples can help models converge quickly and when the models have the capability to recognize the original samples, the ``hard" samples can make them learn more robust features.
Figure~\ref{fig: polciy vs} shows augmented images on the different training phases.
The model gradually receives more adversarial augmented images.
And, these augmented policies learned by \method increase the diversity of training data and highlight the key information of data.

The same analysis of the Reduced SVHN dataset is  presented in Appendix~\ref{app}.
Through analyses of Reduced SVHN and Reduced CIFAR-10 dataset, from an experimental perspective, we illustrate that MADAug consistently provides higher-quality data augmentation strategies for samples, leading to improve more accuracy of the task model across different categories than AdaAug.
From a methodological perspective, we also provide a detailed account of the advantages of MADAug over AdaAug in Appendix~\ref{app}.

\subsection{Computing cost of \method}

To demonstrate the effectiveness of \method, we present a comparison of GPU hours needed to search the augmentation policy and train the task model across different baselines. The results are showed in Table~\ref{tab:hour}. The searching time of our method is regarded as the time to optimize Eq.~\ref{eq:optimaiza}. 
Thus, we do not need extra time to find data augmentation policies.
Our approach is more effective than AutoAugment~\cite{cubuk2018autoaugment}, PBA~\cite{ho2019population}, and AdaAug~\cite{cheung2021AdaAug}.

\begin{table}[h]
	\centering
	\caption{\textbf{Time consumption.} Comparison of computing cost (GPU hours)  in training Wide-ResNet-28-10 on Reduced CIFAR-10 datasets between AutoAugment, PBA, AdaAug, and \method.} %The computing cost (hours)  is estimated on a single GPU.}
\begin{tabular}{lccc}
        \hline
		\multirow{2}{*}{Method} & \multicolumn{3}{c}{Computing Cost} 
       \\ \cline{2-4}
		& \multicolumn{1}{c}{Searching} & \multicolumn{1}{c}{Training} & \multicolumn{1}{c}{Total} \\\hline

  AutoAugment             & 5000         &1.2               & 5001.2  \\
  PBA            & 5         &1.2               & 6.2  \\
 AdaAug            &  2.9        &1.4              & 4.3  \\
\method   & $\sim$0             &1.8            &\textbf{1.8}\\
 \hline
\end{tabular}
\label{tab:hour}
\end{table}

\subsection{Mean and variance of the experiment results}\label{app:var}

Table~\ref{tab:varr} 
represents the average values and variances of these experimental results obtained from 
 multiply trials on different benchmark datasets.
\begin{table}[h]
\centering
\caption{ \textbf{Mean and variance of experiment results.} Test error and  variance (\%) of MADAug on different benchmark datasets with Wide-ResNet-28-10 and ResNet-50.}
\begin{small}
\label{tab:varr}
\resizebox{0.9\columnwidth}{!}{\begin{tabular}{c|c|c|c}
\hline
Dataset&Reduced CIFAR-10 & CIFAR-10 &CIFAR-100  \\
\hline
&12.5$\pm$0.05&2.1$\pm$0.11&16.1$\pm$0.10\\
\hline
Dataset&Reduced SVHN&SVHN&ImageNet\\
\hline
&8.4$\pm$0.09&1.0$\pm$0.10&21.5$\pm$0.15\\

\hline
\end{tabular}}
\end{small}
\end{table}

\section{Ablation study}

\paragraph{Magnitude perturbation.}
Following the AdaAug~\cite{cheung2021AdaAug}, we also add the magnitude perturbation $\delta$ for the augmentation policy.
From  Table~\ref{tab:hyper},  
when the magnitude perturbation of operation is set as 0.3, the performance of the model on the test dataset is best.
And we can conclude that the magnitude perturbation plays a positive effect in improving the generalization of the model.

\paragraph{Number of augmentation operations.}
The number of operations $k$ is arbitrary. Table~\ref{tab:hyper} shows the relationship between the number of operations and the final test error on the Reduced CIFAR-10  with
Wide-ResNet-28-10 model. The number of operations ranges from 1 to 5. When the augmentation operation is chosen as 2,
we have the lowest error rate on the dataset. 
Policies learned by other methods (AutoAugment~\cite{cubuk2018autoaugment}, PBA~\cite{ho2019population}, and AdaAug~\cite{cheung2021AdaAug}) also formulates two augmentation operations.
This phenomenon indicates two augmentation operations'  policies not only increase the diversity and amount of images but also do not make the task model unable to recognize the images due to excessive data augmentations.

\paragraph{Structure of policy network.}
Does the use of a nonlinear projection deliver  better performance? We would add the policy model with the multiple hidden layers and the ReLU activation. Table~\ref{tab:hyper} shows the influence of 
different number $h$ of hidden layers on model performance.
A single linear layer is sufficient for the policy model, without adding extra hidden layers.

\paragraph{Hyperparameter of $\tau$.}
The hyperparameter $\tau$ controls the relationship between the epoch and the number of augmented samples. As is shown in 
Table~\ref{tab:hyper}, the performance of the task model is quite robust to hyperparameter  $\tau \in \{10,20,30,40,50\}$.
For the Reduced CIFAR-10, $\tau$ is optimally set as 40.

\paragraph{Analysis of optimization steps.}
Table~\ref{tab:hyper} illustrates the impact of optimizing data augmentation strategies through different  steps $s$ on the experiment results using the Reduced CIFAR-10. The task model exhibits its highest accuracy when the step size is configured to 1.

\begin{table}[h]
\centering
\caption{\textbf{Ablation study.} Sensitivity analysis of hyperparameter $\delta$, $k$, $h$ $\tau$, and $s$ on Reduced CIFAR-10 (Wide-ResNet-28-10).}
\label{tab:hyper}
\small
\begin{tabular}{c|c|c|c|c|c}
\hline
 $\delta$  &  0     &  0.1  &  0.2 &  0.3   & 0.4    \\
 \hline
 ACC(\%) &86.7 &87.0 &87.3 &\textbf{87.5} &87.2 \\
 \hline
 \hline
 $k$  & 1    &  2  &  3 &  4  & 5  \\
 \hline
 ACC(\%) &86.6&\textbf{87.5}&87.3&86.8&86.2  \\
 \hline
 \hline
 $h$  & 0   &  1  &  2 &  3  & 4  \\
 \hline
 ACC(\%) &\textbf{87.5}&87.2&87.1&86.9&86.7  \\
 \hline
 \hline
 $\tau$  & 10   &  20  &  30 &  40  & 50  \\
 \hline
 ACC(\%) &87.0&87.3&87.4&\textbf{87.5}&87.3  \\
 \hline
 \hline
 $s$  & 1 &2 &5 &10 &30  \\
 \hline
 ACC(\%) &\textbf{87.5} &87.0 &86.6 &86.3 &85.9  \\
 \hline
\end{tabular}
\end{table}

\paragraph{Effect of \trick.}

We investigate the effect of \trick which is introduced in Section~\ref{sec;track}.
We train the Wide-ResNet-28-10  on the Reduced CIFAR-10 and Reduced SVHN datasets without/with this trick across different baselines.
The results are shown in Table~\ref{tab:mm}.
For \method, \trick contributes
to the improvement of accuracy in these datasets.
For other baselines, whether  AutoAugment~\cite{cubuk2018autoaugment} method that applies the same data augmentation policy for the entire dataset, or AdaAug approach that offers different data augmentation policies to different samples, \trick has been found effectively. 

\begin{table}[h]
\centering
\caption{\textbf{Effect of \trick.} Test error (\%) of \method and other baselines without/with \trick.}
\label{tab:mm}
\resizebox{0.9\columnwidth}{!}{\begin{tabular}{c|c|c|c}
\hline
Method & Monotonic curriculum  &Reduced CIFAR-10 &Reduced SVHN \\
\hline
\multirow{2}*{AA}& &14.1 &8.2\\
~&$\checkmark$  &\textbf{13.7}&\textbf{7.8}\\
\hline
\multirow{2}*{AdaAug}& & 15.0&9.1\\
~&$\checkmark$ &\textbf{14.4} &\textbf{8.7}\\
\hline
\multirow{2}*{\method}& & 13.1 &8.9\\
~&$\checkmark$ & \textbf{12.5} &\textbf{8.4}\\
\hline
\end{tabular}}
\end{table}

\paragraph{Strategy of \method.}
\method not only dynamically adjusts the augmentation strategies to minimize the loss of the task model on the validation set which is named a model-adaptive strategy but also provides different data augmentation policies for each sample  called data-adaptive strategy. 
In order to verify the effectiveness of \method, we use one of these two strategies to find the augmentation policies and train the task model to classify the dataset. Table~\ref{tab:model based} shows \method combines these two training strategies well and offers the higher quality of data augmentation policies for the dataset.

\begin{table}[h]
\centering
\caption{\textbf{Effect of model/data-adaptive augmentation strategy.} Test error (\%) of model-adaptive/data-adaptive only \method on two datasets.}

\label{tab:model based}
\resizebox{0.9\columnwidth}{!}{\begin{tabular}{cc|c|c}
\hline
Model-adaptive & Data-adaptive &Reduced CIFAR-10 &Reduced SVHN \\
\hline
$\checkmark$& &14.5 &9.1 \\
&$\checkmark$  &14.0&9.6\\
$\checkmark$&$\checkmark$ & \textbf{12.5}&\textbf{8.3}\\
\hline
\end{tabular}}
\end{table}

\section{Conclusion}
In this paper, we propose a novel and general data augmentation method, \method, which is able to produce instance-adaptive augmentations adaptive to different training stages. Compared to previous methods, \method is featured by a~\trick that progressively increases augmented data and a policy network that generates augmentations optimized to minimize the validation loss of a task model.   
% to search the data augmentation for a dataset called \method to effectively learn model-adaptive augmentation policies. 
\method achieves SOTA performance on several benchmark datasets and its learned augmentation policy network is transferable to unseen tasks and brings more improvement than other augmentations. 
% learned by \method also transfers well to unseen fine-grained datasets. 
We show that \method-augmentations preserve the key information of images and change with the task model in different training stages accordingly. 
Due to its data-and-model-adaptive property, \method has a great potential to improve a rich class of machine learning tasks in different domains.  

%% file: 06-appendix.tex
\appendix
\section{Appendix}

\paragraph{Quality of augmentation policies on Reduced SVHN.}

\begin{figure}[h]
    \centering
    \includegraphics[width=1.0\linewidth]{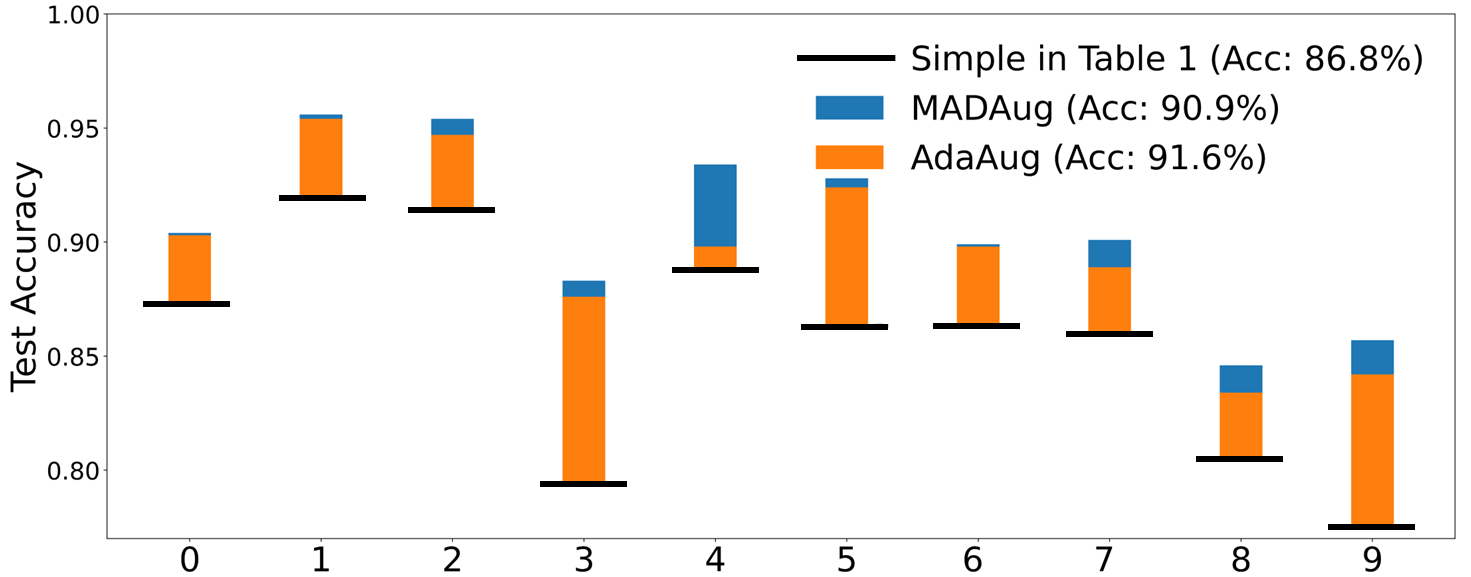}
    \caption{\textbf{MADAug and AdaAug's improvements to different classes on  Reduced SVHN dataset.} 
    % The black line is sample baseline accuracy only with basic augmentation.
    Compared with AdaAug, \method  enhances the test accuracy across various classes, particularly demonstrating a more notable positive impact on same classes such as “4", “7", “8", and “9".
    % considerably increases accuracy in the ``Bird", ``Cat", ``Deer", and ``Dog" classes. Data augmentation learned by AduAug has a negative impact  on the ``Airplane" and ``Automobile" classes.
    }
    \label{fig: per-class accuracy of SVHN}
    %\vspace{-1.0em}
\end{figure}

For Reduced SVHN dataset, we  show data augmentations learned by  MADAug
are superior to AdaAug on the improvement of per-class accuracy, especially in the “4”, “7”, “8”, and “9” classes in Figure~\ref{fig: per-class accuracy of SVHN}.
%In Figure~\ref{fig: per-class accuracy of SVHN}, the augmentation policies learned by \method  are superior to  AdaAug on the improvement of per-class accuracy, especially in the ``4", ``7", ``8", and ``9" classes.
%Figure~\ref{fig: svhn polciy vs} shows some augmented images which are randomly selected from these classes. 
%\begin{figure}[h]
 %   \centering
%    \includegraphics[width=1.0\linewidth]{picture/SVHN sample based.png}
%    \caption{\textbf{MADAug and AdaAug's improvements to different classes on the Reduced SVHN dataset.} 
    % The black line is sample baseline accuracy only with basic augmentation.
  %  Compared with AdaAug, \method consistently increases test accuracy in different classes and has a stronger positive impact on challenging classes like "4", "7", "8", and "9" classes.
    % considerably increases accuracy in the ``Bird", ``Cat", ``Deer", and ``Dog" classes. Data augmentation learned by AduAug has a negative impact  on the ``Airplane" and ``Automobile" classes.
 %   }
 %   \label{fig: per-class accuracy of SVHN}
%\end{figure} 

\begin{figure}[h]
    \centering
    \includegraphics[width=1.0\linewidth]{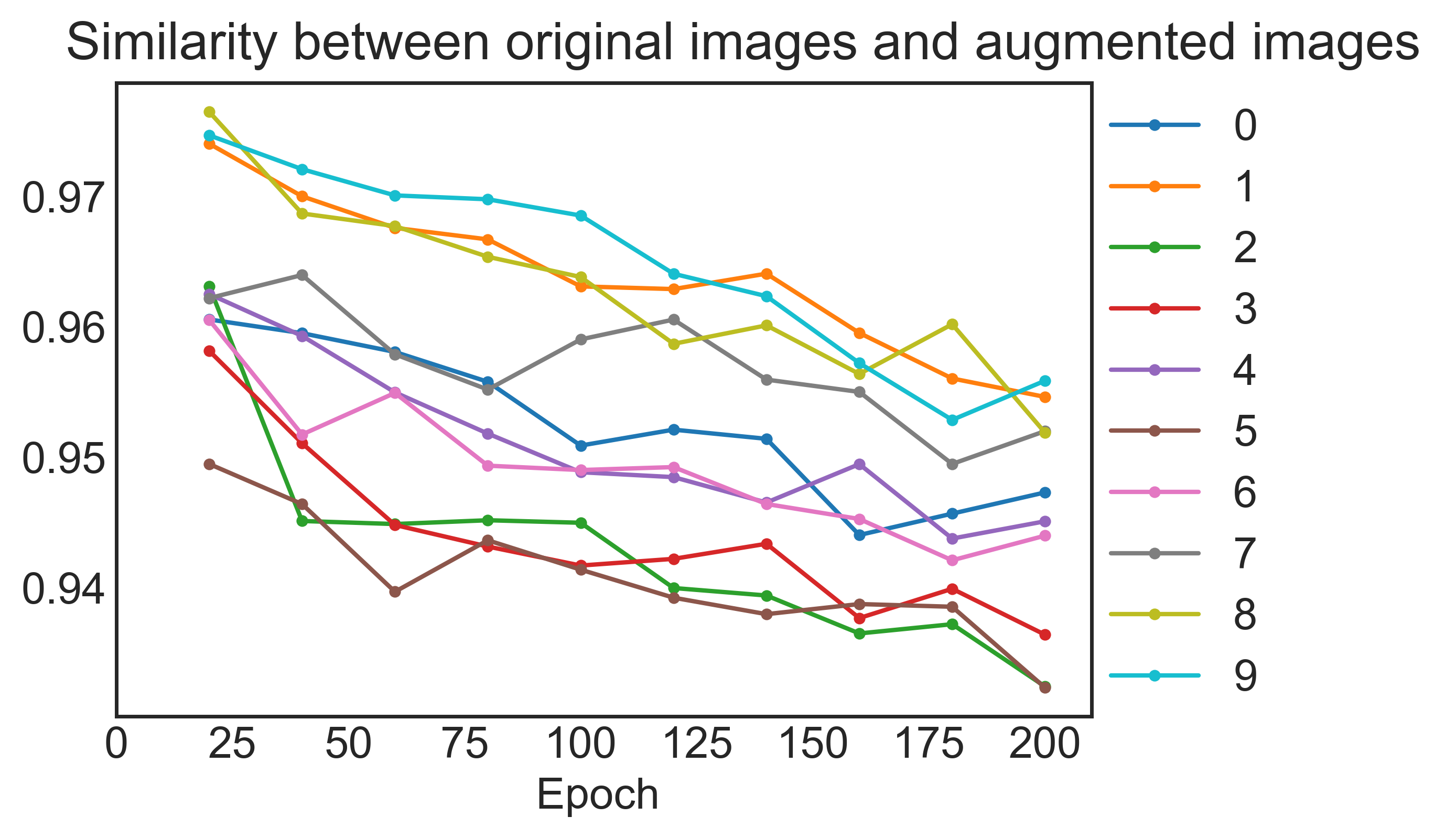}    \caption{\textbf{Similarity between the original images and \method-augmented images at various training phases.} As the training process, \method gradually generates more  ``challenging" augmentation policies for images. }
    \label{fig: svhnsimility}
\end{figure} 

\begin{figure}[h]
    \centering
    \includegraphics[width=1.0\linewidth]{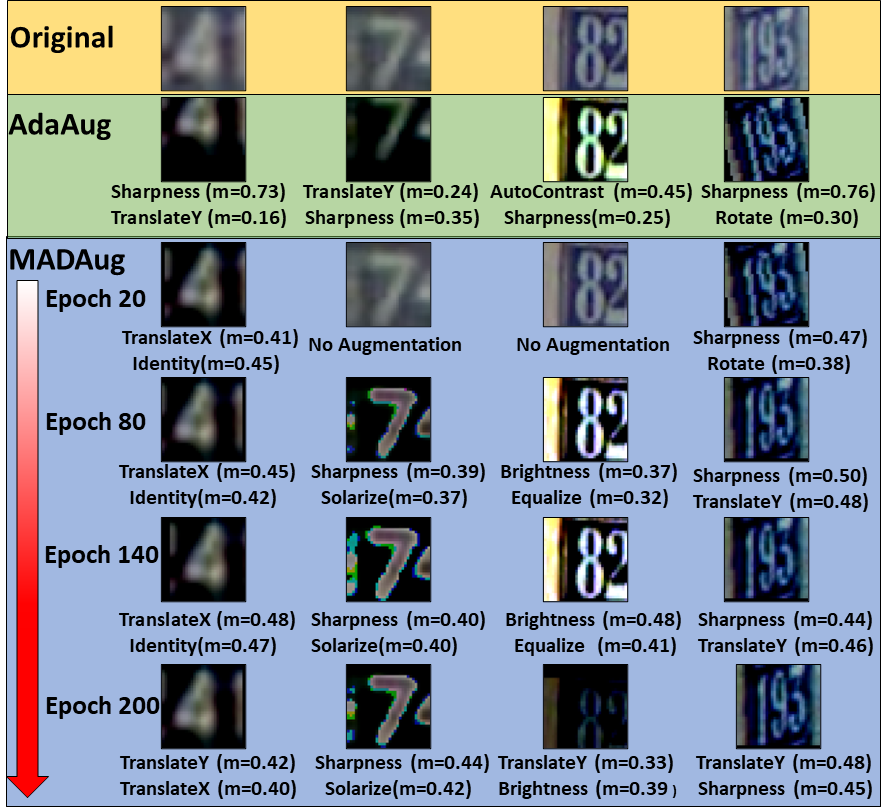}  \caption{ \textbf{Augmentations learned by AdaAug and \method applied to the “4", “7", “8", and “9" class images at
various training epochs  on Reduced
SVHN dataset.} Each augmentation operation is formatted with its name
and magnitude, respectively.%\method introduces less distortion than AdaAug.
    % Data augmentations learned by AdaAug and \method are applied to different class images on the Reduced CIFAR-10 dataset.  Each augmentation is provided with its operation and magnitude respectively.
    % \method produces a wide range of accurate augmented images without distorting their original meanings.
    }
    \label{fig: svhn polciy vs}
    
\end{figure}

AdaAug~\cite{cheung2021AdaAug} provides detailed augmentation policies for models throughout the entire training stage. These  augmentation policies make samples distinguish from the original images, which can potentially hinder model convergence during the early stages of training.
However, \method gradually  applies the data augmentations for samples. 
Figure~\ref{fig: svhnsimility} demonstrates that as the training epoch progresses, more ``adversarial" images generated by \method can be provided for the task model. 
%This matches our motivations.
Figure~\ref{fig: svhn polciy vs} visually illustrates that during the early training phase, the model receives the original images, while as the training progresses, \method learns and applies more “challenging" augmentation policies to augment the images. However, these policies are designed to avoid collapsing the intrinsic meanings of the images and instead, emphasize the crucial information within them.

\paragraph{Advantages of MADAug over AdaAug.}\label{app}

AdaAug employs a two-stage training process, where the first stage involves learning data augmentation strategies for individual samples through alternating ``exploration" and ``exploitation" steps. In the second stage, the learned strategies are fixed while training the task model. However, AdaAug exhibits limitations, including the inability to dynamically update the learned data augmentation strategies based on the performance of the current task model and the suboptimal choice of a two-stage training process.

To address these drawbacks, we propose the MADAug method that overcomes these limitations. In MADAug, we dynamically optimize the data augmentation strategies for each sample by leveraging the current model's performance on the validation set. This facilitates training the model with the most effective data augmentation for the given training stage. We adopt an end-to-end training approach for the task model, differing from AdaAug and AutoAugment, which utilize a two-stage training methodology.

Additionally, we discover that data augmentations do not improve model performance obviously in the early stages of training. Therefore, we use the monotonic curriculum strategy, gradually applying data augmentations to each sample as the training progresses, thereby enhancing the robustness of the task model.

Through the use of MADAug, we demonstrate significant advancements over AdaAug, achieved by its ability to dynamically optimize data augmentation strategies and employ the monotonic curriculum strategy. %Consequently, from Table~\ref{tab:benchmark}, MADAug yields improved performance in image classification tasks.

\paragraph{Model hyperparameters.}
We show the important hyperparameters on different benchmark datasets in Table~\ref{table: hp}. For other details
on the hyperparameters and implementation, we display them in the open
source code.

\begin{table*}[h]
\caption{\textbf{Hyperparameters  on benchmark datasets.} We do not specifically tune these hyperparameters, and all
of these are consistent with PBA and AutoAugment.}
\label{table: hp}
\begin{center}
\begin{small}
% \begin{sc}
\begin{tabular}{cccccc}
\hline
Dataset & Model & Learning Rate & Weight Decay & Batch Size&epoch \\
\hline
CIFAR-10 & Wide-ResNet-40-2    &   0.1 & 5e-4 & 128& 200\\
CIFAR-10 & Wide-ResNet-28-10    &   0.1 & 5e-4 & 128& 200 \\
CIFAR-10 & Shake-Shake (26 2x96d) & 0.01 & 1e-3 & 128& 1,800 \\
CIFAR-10 & PyramidNet+ShakeDrop   & 0.05 & 5e-5 & 64 & 1,800\\
CIFAR-100 & Wide-ResNet-40-2    &   0.1 & 5e-4 & 128& 200 \\
CIFAR-100 & Wide-ResNet-28-10    &   0.1 & 5e-4 & 128& 200 \\
CIFAR-100 & Shake-Shake (26 2x96d) & 0.01 & 2.5e-3 & 128 & 1,800\\
CIFAR-100 & PyramidNet+ShakeDrop   & 0.025 & 5e-4 & 64& 1,800 \\
Reduced CIFAR-10 & Wide-ResNet-28-10    &   0.05 & 5e-3 & 128 & 200\\
Reduced CIFAR-10 & Shake-Shake (26 2x96d) & 0.025 & 2.5e-3 & 128& 1,800 \\
SVHN & Wide-ResNet-28-10 & 0.005 & 1e-3 & 128& 200 \\
SVHN & Shake-Shake (26 2x96d) & 0.01 & 1.5e-4 & 128& 1,800 \\
Reduced SVHN & Wide-ResNet-28-10 & 0.05 & 1e-2 & 128& 200 \\
Reduced SVHN & Shake-Shake (26 2x96d) & 0.025 & 5e-3 & 128& 1,800 \\
ImageNet &ResNet-50 &1.6 &1e-4 &4096& 270\\
\hline
\end{tabular}
% \end{sc}
\end{small}
\end{center}
\vskip -0.1in
\end{table*}